\documentclass[11pt,a4paper]{article}
\usepackage[utf8]{inputenc}
\usepackage[T1]{fontenc}
\usepackage{amsmath}
\usepackage{amssymb}
\usepackage{natbib}
\usepackage{hyperref}
\usepackage{booktabs}
\usepackage{graphicx}
\usepackage{enumitem}
\usepackage[margin=2.5cm]{geometry}
\usepackage{xcolor}
\usepackage{authblk}

\title{Runtime Governance for AI Agents:\\Policies on Paths}

\author[1,2]{Maurits Kaptein\thanks{Corresponding author. Email: \texttt{m.c.kaptein@tue.nl}.}}
\author[2]{Vassilis-Javed Khan}
\author[ ]{Andriy Podstavnychy}

\affil[1]{Eindhoven University of Technology, Eindhoven, The Netherlands}
\affil[2]{Kyvvu B.V., Nijmegen, The Netherlands}
\begin{document}
\maketitle

\begin{abstract}
AI agents---systems that plan, reason, and act using large language models---produce non-deterministic, path-dependent behavior that cannot be fully governed at design time, where with governed we mean striking the right balance between as high as possible successful task completion rate and the legal, data-breach, reputational and other costs associated with running agents.  We argue that the execution path is the central object for effective runtime governance and formalize compliance policies as deterministic functions mapping agent identity, partial path, proposed next action, and organizational state to a policy violation probability. We show that prompt-level instructions (and "system prompts"), and static access control are special cases of this framework: the former shape the distribution over paths without actually evaluating them; the latter evaluates deterministic policies that ignore the path (i.e., these can only account for a specific subset of all possible paths). In our view, runtime evaluation is the general case, and it is necessary for any path-dependent policy. We develop the formal framework for analyzing AI agent governance, present concrete policy examples (inspired by the AI act), discuss a reference implementation, and identify open problems including risk calibration and the limits of enforced compliance.
\end{abstract}

\noindent\textbf{Keywords:} AI agents, runtime governance, compliance policies,
execution paths, organizational risk, EU AI Act

\section{Introduction}\label{sec:intro}

Organizations across industries are deploying AI agents: systems that use large language models to autonomously plan, invoke tools, and take actions with real-world consequences \citep{xi2023agents, wang2024survey, yao2023react}. The attraction is clear---agents can handle complex, multi-step workflows that previously required human labor---and the pace of adoption is rapid. But, governance infrastructure has not kept up. A 2026 KPMG survey of large-enterprise leaders found that 75\% cite security, compliance, and auditability as the most critical requirements for agent deployment, while multi-agent orchestration complexity has become the primary bottleneck as organizations move from pilots to production \citep{kpmg2026pulse}. However, the EU AI Act's provisions for high-risk AI systems --- those making or supporting consequential decisions affecting individuals' rights, safety, or access to services --- take effect in August 2026 \citep{euaiact2024} and effectively demand proper orchestration. The gap between what organizations are deploying and what they can demonstrably govern is, in our assessment, the central obstacle to responsible agent adoption. It is that gap we aim address, at a conceptual level, with this paper.

The governance problem for agents is fundamentally different from that of conventional software or single-query AI systems, and the difference is not merely one of scale. An agent asked to ``prepare a quarterly report'' may read from a CRM, access financial databases, fetch competitor data from the web, generate visualizations, and email the result---a sequence of autonomous decisions, each with its own compliance implications, determined at runtime by a language model whose outputs are stochastic. The same agent on the same task may follow different sequences on different runs --- a fundamental shift from traditional software, where security and control relied largely on predictable, auditable workflows. Hence, the set of possible behaviors is combinatorially large and, for agents with code execution, potentially unbounded. This non-determinism, combined with multi-step tool use and the ability of agents to delegate to other agents, means that the violations organizations care about---data exfiltration, information barrier breaches, unauthorized external communication---are properties of \emph{sequences} of actions, not of individual actions in isolation. A single database read is innocuous; a database read followed by an external email is a potential exfiltration event. No inspection of either step alone reveals the violation.

The objective of governance in this setting is not new: organizations have always sought to maximize the productive output of automated systems while keeping the expected cost of policy violations within acceptable bounds. This is what role-based access control, data loss prevention, and information security frameworks have always aimed to do. What is new is that these mechanisms---designed for deterministic, statically specified systems---cannot express or enforce the path-dependent constraints that agents require. Prompting --- a seemingly popular governance method --- reduces the probability of bad paths but provides no strict enforcement. Access control eliminates action categories (i.e., removes possible agent paths) unconditionally but hardly ever conditions on prior actions which are essential to judge the potential costs. Simply put, neither prompting nor standard access control will reliably detect that an agent has crossed an information barrier by combining the outputs of two individually permitted steps --- and restricting agent behavior post-hoc may not help, as constraints can inadvertently push agents toward alternative paths that were never anticipated.

This paper provides a formal framework for runtime agent governance. We define compliance policies as deterministic functions on execution paths, show that existing governance mechanisms are special cases (or, in the case of prompting, not cases at all), and connect per-step evaluation to an organizational risk objective. The framework is deliberately minimal: it specifies the structure of governance without prescribing specific risk taxonomies or policy languages, which will vary across organizations and evolve as the field matures.

\subsection{Prior work}\label{sec:prior}

The growing recognition that agents require governance beyond what existing mechanisms provide has produced several lines of work. We briefly survey the most relevant contributions and identify the gaps our framework addresses.

\citet{wang2025mi9} introduce MI9, combining an agency-risk index with FSM-based conformance engines for individual agents; AgentSpec \citep{wang2026agentspec} defines a domain-specific language for runtime constraints. Both target individual agents and do not formalize the relationship between per-step evaluation and the organizational risk objective. \citet{gaurav2025gaas} propose Governance-as-a-Service, an external enforcement layer architecturally closest to our approach, but it seems primarily focused on content-level violations (misinformation, hate speech) rather than behavioral trajectory violations. \citet{bhardwaj2026abc} develop Agent Behavioral Contracts with probabilistic compliance guarantees and a Drift Bounds Theorem---the most formally rigorous contribution in this space, complementary to our organizational framing: ABC provides per-agent guarantees; we provide the fleet-level framework within which such guarantees operate.

On the broader landscape, the ArbiterOS paradigm \citep{arbiterOS2025} proposes reframing the LLM as a ``Probabilistic CPU,'' sharing our view that agents are fundamentally different from traditional software. Threat taxonomies from \citet{deng2025agentic_security}, \citet{owasp2025agentic}, and the MAESTRO framework \citep{csamaestro2025} inform our adversarial scenarios. \citet{cihon2025autonomy} propose code-based measurement of agent autonomy levels, providing a complementary design-time assessment that does not address runtime enforcement. \citet{ivliev2025systemic} analyze systemic risk from self-improving agents and argue that static defenses are insufficient against self-modifying systems---a conclusion that aligns with our framework's emphasis on runtime evaluation, though our scope is the governance of current-generation enterprise agents rather than containment of recursively self-improving systems.

What is missing across this body of work is threefold: (i)~a unified model in which existing governance approaches are special cases of a single framework, making precise what each current mechanism can and cannot enforce; (ii)~organizational scope, connecting per-step evaluation to an aggregate risk objective across a fleet of agents; and (iii)~concrete policy specifications tied to regulatory requirements, demonstrating that the framework is not merely theoretical.

\subsection{Aims and outline}\label{sec:aims}

This paper does not present an implementation or experimental results. Its contribution is conceptual: a framework precise enough to guide implementation, general enough to survive changes in the underlying technology, and concrete enough to be immediately useful. Admittedly, this paper is written in a time where new agent paradigms and capabilities are quickly emerging: we do not expect this paper to be the "tell all" of agent governance; we merely strive to provide a meaningful contribution to the literature and help practitioners navigate this space to build agents that are both useful and well-behaved.

Section~\ref{sec:challenge} introduces the governance challenge somewhat informally: we discuss what agents are, what effective governance means, why agents require a different approach, and why current methods fail (or do not address all challenges). This section uses no formal notation and is intended to make the case for runtime governance (as opposed to mere design-time governance) accessible to a broad audience, including policymakers and enterprise architects. Section~\ref{sec:formal} makes the argument more precise: formal definitions of execution paths, the policy function, the Policy Engine and its governance objective, existing approaches as special cases, and a concrete instantiation with worked policy examples. We have tried to capture generically all the cases that we are currently aware of. Next, section~\ref{sec:implementation} discusses how to build a system that realizes this framework, including a reference implementation that illustrates the trade-oﬀs concretely. We discuss potential architectures, practical policy authoring, and the challenges of shared states. Section~\ref{sec:euaiact} reflects on what the framework implies for organizations seeking to comply with the EU AI Act; we hope to provide a meaningful lens. Section~\ref{sec:discussion} summarizes the argument, identifies open problems, and outlines next steps.

\section{The Governance Challenge}\label{sec:challenge}

\subsection{What are agents}\label{sec:whatagents}

For the purposes of this paper, an AI agent is a system that receives a task (prompted or otherwise) described in natural language and autonomously executes a sequence of actions to accomplish the task, where the actions, their order, and their number \emph{are determined at runtime by a language model}. The key components often are: 
\begin{itemize}
\item a language model that decides what to do next, given the task, 
\item the history of prior actions, and any retrieved context; 
\item a set of tools that the agent can invoke---database queries, API calls, web requests, code execution, email, messaging; 
\item "long-term" memory, including conversation history and retrieval-augmented context; and, in many implementations, 
\item internal guardrails---developer-specified checks that the agent is instructed to apply to its own behavior.
\end{itemize}

The behavior of an agent on a task produces an \emph{execution path}: a sequence of discrete steps, each consisting of an action type (language model call, tool invocation, delegation to another agent), an input, and an output. A customer service agent might retrieve a support ticket, look up the customer's account, draft a response, check it against a policy, and send it. A financial analyst agent might query a database, fetch market data from an external API, run a calculation, and email the result. Each of these defines a path---a specific sequence of steps taken on a specific execution.

Two features of this path are consequential for governance. First, the path is not predetermined: the language model decides at each step what to do next, and different runs of the same agent on the same task may produce different paths. We have seen paths on tasks range from a handful of steps, to thousands of steps. Second, the steps in the path interact: the output of an early step (e.g., retrieved data) becomes part of the context for later steps (e.g., what the language model decides to include in an email). This means that the governance implications of a step depend on what happened before it. The objective of governance in this setting is not new: organizations have always sought to maximize the productive output of automated systems, i.e., a system following a path that leads to successful task completion, while keeping the expected cost of policy violations within acceptable bounds. This is what role-based access control, data loss prevention systems, and information security frameworks have always aimed to do. What is new, as we develop in the following sections, is the difficulty of achieving it.

\subsection{Why agents are different}\label{sec:whydifferent}

Five properties make agents qualitatively different from the systems that existing governance mechanisms were designed for, and together they explain why the familiar objective has become substantially harder to achieve.

\begin{enumerate}
\item \emph{Non-determinism.} The same agent given the same task may follow different paths on different runs. This is not a bug---it is constitutive of the flexibility that makes agents useful. But it means that design-time verification of ``the'' behavior is impossible: there is no single behavior to verify.
\item \emph{Dynamic tool use.} Which tools the agent invokes, in what order, and with what arguments are runtime decisions by the language model, not a predetermined sequence specified in code. A traditional workflow automation system calls a fixed sequence of APIs; an agent decides which APIs to call based on what it has observed so far.
\item \emph{Variable-length paths.} Different executions of the same agent on the same task may involve different numbers of steps. The decision surface that governance must cover varies across runs.
\item \emph{Self-modification.} Agents with code execution capabilities can write new functions, modify their own prompts, or create persistent tools at runtime \citep{steinberger2025openclaw}. An agent that writes a helper function to send emails directly---bypassing a governed email tool---has altered its own capabilities in a way that was not anticipated at design time. This is an emerging rather than universal concern: most currently deployed enterprise agents do not have unconstrained code execution. It is, however, a governance-relevant risk for any agent that does.
\item \emph{Multi-agent interaction.} Agents delegate to other agents, receive results from them, and share workspaces. This means that the behavior of one agent can create compliance constraints for another. Consider a financial institution with an advisory-side agent and a trading-side agent separated by an information barrier. If the advisory agent accesses pending deal data and then delegates a task to the trading agent, the trading agent's response may contain deal-adjacent information that the advisory agent now holds alongside restricted data. Neither agent individually violated a rule; the violation is a property of the interaction. Detecting it requires visibility across both agents' paths---which is why governance must be organizational, not per-agent.
\end{enumerate}

Sadly, in the sense that it complicates our governance of agents, these properties all interact. An agent that non-deterministically decides to fetch external data (dynamic tool use), then writes a script to process it (self-modification), then delegates to another agent for a summary (multi-agent interaction) has produced a path whose governance implications could not have been specified at design time, whose length was not predictable, and whose compliance depends on the full sequence of steps including interactions with other agents. This is the governance challenge in fairly concrete terms.

We use three concrete scenarios that we return to later in the paper to make things concrete:
\begin{enumerate}
\item A customer service agent retrieves a support ticket containing injected instructions \citep{greshake2023indirect, perez2022red} and then acts on those instructions, disclosing account data. Neither step---retrieving a ticket, answering a query---is inherently violating; the violation is in the sequence. 
\item An agent preparing a report reads from a CRM, accesses financial data, fetches competitor pricing from the web, and emails the report externally. The external email may violate policy---but only because of what data was accessed in prior steps. 
\item Two agents, each acting within its permissions, produce a combined path that violates a cross-organizational constraint. In the formalization of Section~\ref{sec:formal}, each of these scenarios becomes a concrete policy (the information barrier scenario described above).
\end{enumerate}

A natural---and common ---response to these scenarios is to require human approval before consequential actions. This is valuable---and the framework we present in Section \ref{sec:formal} accommodates it---but it does not dissolve the governance problem. Human approval addresses the question of whether a \emph{specific} action is acceptable at the moment it is proposed; it does not address whether the path leading to that point already contains a violation, whether the volume of approval requests is manageable, or whether a reviewer presented with a wall of context under time pressure is providing meaningful oversight or a rubber stamp. Human approval will not be fault proof and should simply be regarded another step in the potentially risky agent's execution path. The potentially more interesting question is not whether human approval solve our governance problem, but rather when, on what information, and decided by what mechanism, human approval should be one of the steps in the agent's task execution. This is exactly the types of questions a runtime governance layer should aim to answer.

\subsection{Why current methods fall short}\label{sec:boundary}

We now examine five categories of existing, common, governance mechanisms and, for each, identify what it can and cannot do in the context of agents. We would argue that all are useful, none are complete.

\begin{itemize}
\item \emph{Prompt-level control.} System prompts, injected instructions, "system prompts" and in-context examples shape the agent's behavior by modifying its input. An agent told ``never send emails containing customer data without approval'' is statistically less likely to do so. But prompt compliance is probabilistic, not guaranteed: the agent may ignore, misinterpret, or be manipulated into overriding its instructions, as the prompt injection scenario illustrates. More fundamentally, prompting cannot \emph{enforce} anything---there is no mechanism that prevents an action if the agent decides to take it. Prompting reduces the probability of policy-violating paths; it provides no enforcement when violations occur.
\item \emph{Access control (RBAC, IAM, tool whitelisting).} Role-based access control and tool whitelisting restrict which actions an agent can take, unconditionally. An agent without email permissions cannot send emails; an agent without database access cannot query databases. This is genuine enforcement---the action is categorically prevented and hence any possible task execution paths containing that action node are cut. But it is (at least in most implementations) context-free: access control does not know, and cannot condition on, what the agent has done before. An agent with both ``database read'' and ``email send'' permissions has everything it needs to exfiltrate data, and access control cannot distinguish the innocuous use of these permissions from the problematic one. Access control prevents action \emph{categories}; it cannot prevent action \emph{sequences}.
\item \emph{Agent-level guardrails.} Many agent frameworks include developer-specified checks: output filters, content classifiers, self-critique steps. These operate under the agent's own authority---they are part of the agent's code, executed within the agent's process. For agents with code execution capabilities, this is a fundamental limitation: an agent that can write code can, in principle, write code that bypasses its own guardrails. However, even without self-modification, guardrails embedded within the agent cannot be audited, updated, or enforced independently of the agent. They are not governance; they are (potentially useful) self-regulation.
\item \emph{Content-level filtering.} External content filters \citep{gaurav2025gaas} can intercept and evaluate the content of individual steps---checking for PII, hate speech, or misinformation. This is valuable but addresses a different problem: content violations at individual steps, not behavioral trajectory violations across steps. A content filter can detect that an email contains a social security number; it cannot detect that the email was sent after the agent read confidential deal data from a restricted database. The true violation is in the trajectory, not the content and hence content level filtering will not suffice.
\item \emph{Human approval.} Requiring a human to approve consequential actions is often treated as a catch-all solution, and it is genuinely useful: a human who sees the accumulated context of an agent's path can exercise judgment that no policy function can replicate. But human approval is not a governance mechanism in its own right---it is an action that a governance mechanism can invoke. The policy that decides \emph{when} to request approval, what context to surface to the reviewer, and what the approval unlocks in subsequent steps is where the governance work lives. Approval without that surrounding structure is unscalable (a fleet of agents generates more requests than humans can meaningfully review), incomplete (the path leading to the approval point may already contain a violation), and gameable (an agent can be manipulated into reaching the approval gate via a sequence of individually innocuous steps that together constitute a violation). Used well---invoked selectively by a policy that has already evaluated the path, and presented to a reviewer with the relevant context made explicit---human approval is a powerful component of the governance system. It should however not be mistaken for the system itself.
\end{itemize}

Taken together, these mechanisms form a layered defense that every organization deploying agents should maintain: prompting reduces the base rate of violations; access control eliminates entire action categories; guardrails provide agent-level checks; content filtering catches per-step content issues. Each is valuable. None however can express or enforce a policy that depends on the sequence of prior actions. For that, a different mechanism is needed: one that evaluates the proposed next action in the context of the full execution path, operates externally to the agent, and applies uniformly across all agents in the organization. This is the runtime governance framework we formalize in Section~\ref{sec:formal}.

\section{A Formal Framework for Agent Governance}\label{sec:formal}

We proceed as follows. We first define the execution path of an individual agent executing a task---the object we govern---and the step types that constitute it. We then introduce the policy function, which maps a partial path and a proposed next action to a violation probability. Here we intentionally refrain from discussing policy implementations in detail; we, for the time being, merely assert that such policies can meaningfully be constructed. Next, we introduce a Policy Engine, the organizational component that evaluates policies across a fleet of agents and keeps expected violations within an acceptable bound. We close by showing how prompting and access control relate to this framework

Throughout, we distinguish the \emph{ideal formalization}---what a complete governance system should do in principle---from more \emph{workable instantiations} that approximate the ideal under practical constraints. The gap between the two is not a weakness of the framework; it is the precise statement of the open organizational and engineering problems we face when governing agents.

\subsection{The Execution Path}\label{sec:paths}

An \textbf{agent} $A$ is a computational entity with a persistent identity and a registered purpose. For the purposes of the formal framework, $A$ is simply an identifier; Section~\ref{sec:instantiation} discusses what metadata a concrete system attaches to it.

The execution of a task by agent $A$ produces an \textbf{execution path} $P = (s_1, s_2, \dots, s_n)$, a finite sequence of discrete steps. Each step is a triple $s_i = (\tau_i, d_{\mathrm{in},i}, d_{\mathrm{out},i})$, where $\tau_i$ is the step type, $d_{\mathrm{in},i}$ is the input data to that step, and $d_{\mathrm{out},i}$ is the observed output. We distinguish three step types, each presenting a qualitatively different governance concern:

\begin{enumerate}
\item A \emph{stochastic} step is a call to a language model: its output is non-deterministic, conditioned on the full input context, which may have been altered by prior steps---for example, by retrieved content containing injected instructions. 
\item A \emph{deterministic} step is a call to an external tool or system according to a defined schema---database queries, API calls, memory reads and writes---where the governance concern is what data is accessed, modified, or transmitted. 
\item A \emph{composite} step is a delegation to another agent, which will itself produce a sub-path $P'$; this type combines the concerns of the previous two and introduces the additional question of how governance of the sub-path propagates to the primary path, a question we revisit in Section~\ref{sec:open}. 
\end{enumerate}

This taxonomy is motivated by governance, not by implementation: a concrete system will likely subdivide these categories further when policies require finer distinctions.

One might expect a fourth type for human inputs---approval responses or corrections provided by a person during execution. We do not treat this as a primitive type because a human input is, from a governance perspective, either a deterministic step (a structured approval with a defined schema) or a stochastic step (free-form input whose content is unpredictable). Its governance significance lies not in constituting a separate type but in its presence in $P_i$, which approval-gating policies can verify.

What constitutes a step is a design choice of the implementer. A coarser granularity reduces the precision at which policies can intervene; we assume throughout that granularity is fine enough that the governance-relevant decisions---tool invocations, external communications, data accesses, delegations---are individually visible. Informally, the space of all possible executions of an agent on a task can be thought of as an implicit graph, with paths as walks from the initial state to a terminal outcome; the aim is to keep agents on paths that are both successful and low-cost. Policies are functions on observed (and prospective) partial paths, not properties of the graph itself. For realistic agents the true graph is unknown, potentially infinite, and changes at runtime. We assume however that each execution path terminates in a \textbf{terminal state}: either \emph{success} ($\top$), meaning the task completed, or \emph{failure} ($\bot$), meaning the task did not---whether due to natural task failure, human intervention, or automatic policy-engine-induced termination. We associate with each completed task a \textbf{utility} $u(P) \geq 0$, where higher values indicate greater task value; the specific utility function is task-dependent and left to the instantiation.\footnote{A natural baseline is $u(P) = \mathbf{1}[\text{terminal state} = \top]$, i.e.\ binary success, which is sufficient to make the governance objective well-defined and to expose the trivial-block pathology.}

\subsection{The Policy Function}\label{sec:policy}

At step $i$, the partial path $P_i = (s_1, \dots, s_i)$ records what the agent has done so far. We subsequently introduce an agent's intended next action $s^* = (\tau^*, d^*_{\mathrm{in}})$. Note that the output $d^*_{\mathrm{out}}$ of this next action is unknown, since $s^*$ is what the agent intends to do, not what it has done.

In the ideal framework we intend to sketch, we believe the governance system should evaluate $s^*$ \emph{before} it executes, using the full partial path $P_i$ as context. This prospective evaluation is what makes preemptive enforcement possible: a violation can be prevented rather than merely recorded. The alternative---evaluating solely $s_i$ after execution---is retrospective and can only detect and record violations, not prevent them.

An organization's agent \emph{governance} is a set of policies $\mathcal{J}$. Each policy $j \in \mathcal{J}$ is a \emph{deterministic} function
\[
  \pi_j(A,\; P_i,\; s^*,\; \Sigma) \;\longrightarrow\; [0, 1],
\]
whose output is the \textbf{probability that executing $s^*$ constitutes a violation of policy $j$}, given agent $A$, partial path $P_i$, and shared governance state $\Sigma$. 

The input $A$ carries whatever registered metadata the organization associates with the agent. The input $P_i$ records completed steps; in practice, most policies would compress this into a small state vector updated incrementally rather than re-examining the full sequence at each step. The input $s^*$ is the proposed action, consisting of its type $\tau^*$ and input $d^*_{\mathrm{in}}$. The input $\Sigma$ represents the \textbf{shared governance state} maintained by the Policy Engine across all agents: it captures governance-relevant facts that no single agent's path contains, such as which agents have accessed which data categories or which information barriers have been activated. Most policies in practice likely ignore $\Sigma$; policies encoding cross-agent constraints however require it.

In our view, the policy function has to be deterministic: identical inputs always produce identical outputs. This is a deliberate design constraint. If policy evaluation were non-deterministic, the same execution path could produce different enforcement decisions on different runs, making audit logs unverifiable and compliance guarantees impossible to state. Determinism is also what makes an audit record meaningful: given $P_i$, $s^*$, and the value of $\Sigma$ at evaluation time, any auditor can reproduce the scores that led to an enforcement decision.\footnote{Non-deterministic policy functions seem "all the rage"; i.e., we use an agent to evaluate the behavior of another agent. This is easy to incorporate in our framework: the "policy agent" call is simply a step in $P$.} 

It is worth noting that the input space of $\pi_j$ is very large---$P_i$ alone may be arbitrarily long---so in practice every policy maps large regions of the input space to the same output. For the same task, different executions may produce different $P_i$ because stochastic steps are non-deterministic, and hence different policy outputs. The apparent variability of policy outputs across runs reflects the stochasticity of the agent, not of the policy.

\subsection{The Policy Engine and Governance Objective}\label{sec:engine}

Policies are functions; something must evaluate them, act on the results, and maintain the shared state they depend on. We call this the \textbf{Policy Engine}: the organizational component responsible for intercepting proposed actions, evaluating all applicable policies, maintaining $\Sigma$, and issuing interventions.\footnote{A complete implementation also maintains an audit trail. From the framework's perspective this is simply the stored sequence of tuples $(A, P_i, \Sigma, s^*, v_i, \delta(v_i))$ at each step---a natural byproduct of runtime evaluation rather than a separate concern.}

Given a proposed action $s^*$ at step $i$, the Policy Engine evaluates all active policies and combines their outputs into a \textbf{step-level violation score},
\[
  v_i \;=\; 1 - \prod_{j \in \mathcal{J}}\bigl(1 - \pi_j(A, P_i, s^*, \Sigma)\bigr),
\]
the probability that at least one policy is violated by executing $s^*$. Because each $\pi_j$ conditions on the full partial path $P_i$, $v_i$ is already path-level: it reflects not just the proposed action but the history of steps that led to it. This quantity is bounded in $[0,1]$ regardless of the number of policies and does not grow artificially with $|\mathcal{J}|$.

The quantity of interest per task is the \textbf{terminal violation score} $v_T$, the step-level score at the terminal step of the task. The \textbf{fleet-level governance objective} is a constrained optimisation: maximise expected task utility across the fleet while keeping the expected terminal violation score within an acceptable bound,
\[
  \max \; \mathbb{E}\!\left[\sum_{a \in \mathcal{A}(t)} \sum_{\text{tasks of } a} u^{(a)}\right]
  \quad \text{subject to} \quad
  \mathbb{E}\!\left[\sum_{a \in \mathcal{A}(t)} \sum_{\text{tasks of } a} v_T^{(a)}\right] \;\leq\; B,
\]
where $B$ is the organization's risk budget. This formulation makes the governance tradeoff explicit: a policy engine that simply blocks all tasks achieves $\mathbb{E}[\sum v_T] = 0$ trivially, but at the cost of $\mathbb{E}[\sum u] = 0$. The budget $B$ is directly interpretable and can be monitored in real time: $B = 0.1$ means the organization tolerates an expected $0.1$ policy-violating tasks completing at any moment, a quantity a risk officer can calibrate against regulatory exposure.\footnote{It is trivial to include policy violation specific costs into this framework; we intentionally try to forward only the core parts we deem necessary for a richer understanding.}

Finally, the Policy Engine applies a decision function $\delta$ that maps the step-level violation score $v_i$ to an \textbf{intervention}. An intervention either terminates the task---moving it to a failure terminal state $\bot$ and thereby capping $v_T = v_i$---or modifies the path in a way that allows execution to continue with reduced violation risk. The latter includes any action that results in a new step being appended to $P_i$ before the agent proceeds, since subsequent $\pi_j$ evaluations condition on the updated $P_i$ and may therefore return materially lower scores. Crucially, $\delta$ faces a genuine tradeoff: terminating a task at step $i$ caps $v_T$ but reduces $u(P)$; the organization must therefore calibrate $\delta$ so that the fleet-level objective is maintained while aggregate utility remains acceptably high.\footnote{Concrete interventions---blocking, steering, requesting human approval---are implementation choices; we discuss them in Section~\ref{sec:implementation}.}

\subsection{Existing Approaches as Special Cases}\label{sec:special}

Having defined the policy function and the Policy Engine, we can locate existing governance mechanisms precisely within the framework.

Prompt-level control---system prompts, injected instructions, in-context examples---does not instantiate $\pi_j$ at all. Prompts modify the agent's input, shifting the distribution over future paths towards (hopefully) those with lower $\pi_j$ values. An agent instructed \emph{never send emails without approval} will produce proposed actions $s^*$ for which approval-related policies return lower violation probabilities. But this is a shift in distribution over possible execution paths, not an enforcement: there is no mechanism that prevents a high-probability action if the agent proposes one. Prompting makes costly paths less likely; it does not remove them.

Access control---role-based permissions, tool whitelisting, API authentication---implements a degenerate case of $\pi_j$ that uses (often) only $A$ and the proposed action type $\tau^*$:
\[
  \pi_{\mathrm{access}}(A, P_i, s^*, \Sigma) \;=\;
  \begin{cases}
    0 & \text{if } \tau^* \in \mathrm{Allowed}(A), \\
    1 & \text{otherwise.}
  \end{cases}
\]
The partial path $P_i$, the proposed input $d^*_{\mathrm{in}}$, and the shared state $\Sigma$ are all ignored; the output is binary. Access control eliminates action categories unconditionally but cannot express context-dependent constraints. It is a special case of $\pi_j$ with $P_i$, $d^*_{\mathrm{in}}$, and $\Sigma$ fixed to null.

Runtime evaluation instantiates the full $\pi_j(A, P_i, s^*, \Sigma)$, using all four inputs, and is the general case. Access control is the special case where path and input are ignored; prompting is not a case at all. Any policy whose violation condition depends on what happened in prior steps can only be enforced at runtime.

Finally, note that compliance may be \emph{voluntary}---the agent calls the Policy Engine of its own accord and may skip the call---or \emph{enforced}---the Policy Engine is architecturally interposed so that no proposed action executes without passing through $\pi_j$. Only enforced compliance provides verifiable governance guarantees. Voluntary compliance shifts behavior in the right direction; it should not be mistaken for governance.

\subsection{Concrete Instantiation}\label{sec:instantiation}

The framework above is abstractly stated at the level of ideal objects. We now show how each abstract component maps to a concrete choice by working through the three scenarios introduced in Section~\ref{sec:whydifferent}.

To this end we assume that in a concrete system, the agent identity $A$ carries a metadata record $\mathcal{M}_A$ specifying registered purpose, risk classification, owner, tool configuration, and a cryptographic hash $h(A)$ of the agent's definition at registration time. The hash enables detection of self-modification: at task start, the Policy Engine recomputes the hash of the running definition and compares it to $h(A)$. Agent definitions evolve---prompts are revised, tools added---and each change constitutes a re-registration producing a new hash and retiring the old version. The shared governance state $\Sigma$ is instantiated as a ledger maintained by the Policy Engine recording, for each active task, the maximum sensitivity level of data accessed, flags for information-barrier tags that have been activated, and cross-agent delegation records; it is updated each time a deterministic or composite step completes.

With these instantiations in place, the following policies each correspond to one of the three scenarios from Section~\ref{sec:whydifferent}:

\begin{itemize}
\item \textit{Agent integrity}: at task start, the Policy Engine compares the hash of the running definition to $h(A)$, returning $1$ if they differ and $0$ otherwise; depends only on $A$ and can be evaluated before any step executes. 
\item \textit{Purpose and documentation}: returns $1$ if required fields (purpose, risk classification, owner) are absent from $\mathcal{M}_A$; depends only on $A$; evaluated at deployment time. 
\item \textit{PII predecessor requirement}: returns $1$ if $s^*$ would access personal data and no \texttt{PII\_Check} step appears in $P_i$; addresses the prompt-injection scenario by requiring that a classification step precede any action on personal data. 
\item \textit{Approval before external actions}: for high-risk agents, returns $1$ if $s^*$ is an external action and no \texttt{Human\_Approval} step appears in $P_i$; addresses emergent tool chaining by blocking the terminal external send until approval is recorded. 
\item \textit{Data exfiltration prevention}: assumes deterministic steps are labeled with a sensitivity level $\sigma \in [0, \sigma_{\mathrm{ceiling}}]$ written into $\Sigma$ on completion; returns $\sigma_{\max} / \sigma_{\mathrm{ceiling}}$ if $s^*$ sends data externally, and $0$ otherwise---a graduated policy whose score reflects what has been touched in prior steps. \item \textit{Information barrier}: if $\Sigma$ records that $A$ has accessed data tagged as one side of a named barrier and $s^*$ involves the other side, returns $1$; this policy cannot be evaluated from $A$'s path alone, making it the clearest illustration of why $\Sigma$ is necessary. 
\item \textit{Execution bounds}: returns a score increasing linearly with $|P_i|$, reaching $1$ at a configured maximum. 
\item \textit{Time restriction}: returns $1$ if the current time falls outside permitted operating hours for the agent's risk classification; depends only on $A$.
\end{itemize}

Table~\ref{tab:policies} summarizes which inputs each policy uses, whether its output is binary or graduated, and whether it can be evaluated before the task starts. The pre-task column points to a practically important observation: all policies that depend only on $A$ can be evaluated once at deployment or task start, without per-step interception, reducing runtime overhead to those policies that genuinely require $P_i$, $s^*$, or $\Sigma$.

\begin{table}[h]
\centering
\caption{Concrete policy examples, their inputs, output type, and whether they
can be evaluated before the task starts.}
\label{tab:policies}
\begin{tabular}{lcccccl}
\toprule
Policy & $A$ & $P_i$ & $s^*$ & $\Sigma$ & Output & Pre-task \\
\midrule
Agent integrity       & \checkmark &            &            &            & binary    & yes \\
Documentation         & \checkmark &            &            &            & binary    & yes \\
PII predecessor       &            & \checkmark & \checkmark &            & binary    & no  \\
Approval required     & \checkmark & \checkmark & \checkmark &            & binary    & no  \\
Data exfiltration     &            & \checkmark & \checkmark & \checkmark & graduated & no  \\
Information barrier   & \checkmark & \checkmark & \checkmark & \checkmark & binary    & no  \\
Execution bounds      &            & \checkmark &            &            & graduated & no  \\
Time restriction      & \checkmark &            &            &            & binary    & yes \\
\bottomrule
\end{tabular}
\end{table}

The table also makes the paper's central argument concrete. Access control corresponds to policies that use only $A$ and $\tau^*$---the path, proposed input, and shared state are ignored, and the output is binary. Runtime evaluation is required for every other row. No amount of prompting or access restriction can enforce a policy whose violation condition depends on what happened in prior steps.

\section{Implementation}\label{sec:implementation}

The framework of Section~\ref{sec:formal} describes what a governance system should do in principle. This section discusses in more detail what it takes to realize that in practice. We do not present a complete system; the contribution of this paper is intentionally conceptual. Instead, we discuss the key architectural choices that determine how closely an implementation can approach the ideal, the authoring of policies, and a reference implementation that illustrates the tradeoffs concretely.

\subsection{Deployment modes}\label{sec:deployment}

The most consequential implementation decision is whether the Policy Engine evaluates proposed actions \emph{before} or \emph{after} they execute. This single distinction determines what the engine can prevent versus what it can only detect, and it maps directly onto the prospective/retrospective already hinted at in Section~\ref{sec:policy}. In \emph{prospective mode}, the Policy Engine intercepts each proposed action $s^*$ before the underlying call is made, evaluates all applicable policies, computes $v_i$, and applies $\delta$ before execution proceeds. This is enforced compliance: violations can be prevented, not merely recorded. Prospective interception can be achieved at different layers of the stack---wrapping tool APIs, instrumenting the agent framework, or interposing at the model endpoint---but these are engineering choices within the mode, not governance-relevant distinctions. What matters is that no proposed action executes without passing through $\pi_j$. In \emph{retrospective mode}, the Policy Engine receives execution logs after steps complete and evaluates policies post-hoc. No interception occurs; $\delta$ can flag violations and generate alerts but cannot prevent actions. This mode is purely detective: it supports audit, agent profiling, and after-the-fact escalation, but it cannot satisfy regulatory requirements that demand prevention. It is, however, the only mode available when the organization does not control the agent's execution environment---for example, when using third-party agent services that expose logs but not interception hooks.

The two modes are not mutually exclusive: a prospective engine also produces an audit trail, and a retrospective engine can feed its findings back into the registration-phase policy evaluations that gate future task starts. But they are strictly ordered by governance strength: prospective mode is the target; retrospective mode is a fallback when interception is not available.

A practical caveat applies to both modes. As noted in Section~\ref{sec:whydifferent}, agents with code execution capabilities can in principle generate actions that bypass whatever interception layer is in place---writing code that calls tool APIs directly, spawning subprocesses, or modifying their own execution environment. Prospective interception is enforced under the assumption that the agent operates within the governed execution environment; it is an architectural assumption, not a proven invariant. The completeness of enforced compliance under self-modification is an open problem (see Section~\ref{sec:open}).

\subsection{System architecture}\label{sec:sysarch}

Regardless of deployment mode, the Policy Engine operates in two phases that correspond directly to the pre-task/runtime distinction established in Table~\ref{tab:policies}.

In the \emph{registration phase}, when an agent is deployed or re-registered, the Policy Engine evaluates all policies that depend only on $A$. An agent that fails these checks---because it is undocumented, its definition hash has changed, or it is scheduled outside permitted hours---is rejected before any task begins. This eliminates an entire class of violations without per-step overhead and provides a natural enforcement point for deployment-time organizational requirements.

In the \emph{per-step phase}, the Policy Engine intercepts each proposed action $s^*$, evaluates all remaining policies, computes $v_i$, applies $\delta$, and records the step to the audit trail. The key to making this tractable is that most policies do not require re-examination of the full path $P_i$ at each step. Instead, the Policy Engine maintains a compact \emph{governance state vector} per task---a small set of incrementally updated values such as the maximum data sensitivity level encountered, a boolean recording whether approval has occurred, and the current step count. Updating this vector at each step is constant-time; evaluating most policies against it is equally cheap. The state vector is a sufficient statistic for each policy: rather than conditioning on the full path, each policy conditions on the state vector, which captures everything it needs from prior steps. This is exact for all policies in Table~\ref{tab:policies} and for the large majority of practically relevant policies.

Computing $v_i = 1 - \prod_j (1 - \pi_j(\cdot))$ across a policy set $\mathcal{J}$ is linear in $|\mathcal{J}|$ per step. Policies that depend only on $A$ and $\tau^*$ can be short-circuited before the state vector is consulted, and independent policies can be evaluated in parallel. For typical policy sets of tens to low hundreds of policies, per-step evaluation adds modest overhead relative to the language model inference that often dominates agent execution time.

Note that audit logging is a natural byproduct of the per-step phase: the full state tuple $(A, P_i, \Sigma, s^*, v_i, \delta(v_i))$ at each step is already available and can be persisted. The audit trail is itself a high-value asset requiring appropriate access controls and retention policies; see Section~\ref{sec:challenges}.

\subsection{Policy authoring}\label{sec:authoring}

In practice, the large majority of organizationally relevant policies are binary threshold rules on path state: has a particular step type appeared, has a sensitivity level been exceeded, has the step count reached a limit. These are cheap to evaluate, straightforward to audit, and simple to express. The graduated policies in Table~\ref{tab:policies} are the exception rather than the rule, and even these reduce to simple arithmetic on the state vector.

This observation motivates a template-based approach to policy authoring. Rather than expecting compliance officers to write policy functions from scratch, organizations can maintain a library of parameterized templates covering the common governance patterns: approval-gating (which action types require prior approval for which agent risk classifications), data sensitivity thresholds (at what sensitivity level does an external action require intervention), information barrier enforcement (which data categories are barrier-separated), and execution bounds (maximum step counts by agent type). Authoring a policy becomes a matter of selecting a template and configuring its parameters---a task that requires governance judgment but not programming.

Testing policies before deployment is non-trivial. A policy that is too permissive fails to catch violations; one that is too restrictive blocks legitimate task paths and reduces agent utility. Policies also interact via the composition rule $v_i = 1 - \prod_j(1 - \pi_j)$: a new policy that individually appears reasonable can, combined with existing policies, push $v_i$ high enough to trigger interventions on most useful paths. The recommended practice is to deploy new policies in flag-only mode first---computing $v_i$ and logging the results without acting on them---and to validate against a representative sample of execution traces before enabling enforcement. Policy changes take effect across all agents immediately and should be versioned: the audit trail records the active policy version at each step, so that enforcement decisions can be reproduced and reviewed against the policy set that was in force at the time.

\subsection{Concrete intervention outcomes}\label{sec:interventions}

The decision function $\delta$ maps $v_i$ to an intervention at each step. In practice, implementations typically realize this as one of three concrete outcomes. \textsc{Pass}: the proposed action executes and the agent continues unmodified. \textsc{Steer}: execution pauses; the Policy Engine may inject a compliance hint into the agent's context, request human approval, or alert a responsible person, with $P_i$ persisted and execution resuming from the stored state when resolved. \textsc{Block}: the proposed action is prevented, the task terminates at a failure state $\bot$, and the incident is escalated. The thresholds at which $\delta$ produces each outcome are organizational choices, calibrated so that the fleet-level objective is maintained while aggregate utility remains acceptably high.

\subsection{A reference implementation}\label{sec:kyvvu}

To illustrate how the framework translates into a working system, we describe the implementation developed by Kyvvu B.V., documentation for which is available at \url{https://docs.kyvvu.com} (but the platform is in active development). We describe it not as a definitive realization of the framework but as a concrete instantiation that makes the tradeoffs visible.

Kyvvu's Policy Engine operates in prospective mode: agents call the engine before each step, passing the proposed action and receiving a governance decision before proceeding. The engine integrates with LangChain and LangGraph agent frameworks, and with Microsoft Copilot Studio agents covering a large section of current enterprise agent deployments. 

The registration phase evaluates agent-level policies at task start, rejecting agents that fail documentation, integrity, or scheduling checks before any step executes. The per-step phase maintains a governance state vector per task and evaluates the active policy set at each proposed action, producing one of the three intervention outcomes described in Section~\ref{sec:interventions}. Currently, policy scores are treated as severity indicators rather than calibrated probabilities---the outputs of $\pi_j$ reflect the seriousness of the concern but have not been empirically calibrated against actual violation rates, which requires operational data at scale that is not yet available. The shared governance state $\Sigma$ is maintained per-organization with eventual consistency across concurrent agent executions, which is sufficient for information barrier policies under normal operating conditions but may miss violations in high-concurrency edge cases.

What the implementation does address, beyond most current alternatives, is the combination of prospective enforcement, path-level policy evaluation, and organizational scope: policies condition on the full governance state vector rather than just the proposed action type, the engine observes all agents in the organization enabling information barrier enforcement, and the audit trail records the complete state tuple alongside policy scores and decisions at each step. Whether this is sufficient for any given regulatory context depends on the specific requirements and the calibration of $\delta$ against the organization's risk budget $B$.

\subsection{Challenges}\label{sec:challenges}

Several challenges arise in any implementation of the framework and are worth stating explicitly, both as guidance for practitioners and as a record of what the framework does not resolve:

\begin{enumerate}
\item \emph{Fail-closed behavior.} The Policy Engine is a single point of failure in the governance architecture. If it becomes unavailable, the organization must choose between blocking all agent execution (fail-closed) or allowing agents to proceed ungoverned (fail-open). Fail-closed is the only acceptable default for any deployment where governance guarantees are a regulatory requirement. Production deployments require redundancy, health monitoring, and timeouts that default to \textsc{Block}.
\item \emph{Shared state consistency.} The governance state $\Sigma$ is updated concurrently by multiple agents and must be read consistently during policy evaluation. For most policies, eventual consistency is sufficient. For the most sensitive barrier policies, strict consistency may be required, introducing coordination overhead. The state itself is small---a set of tags and counters, not full path records---so the consistency problem is manageable, but it must be designed for explicitly.
\item \emph{Audit trail privacy.} The audit trail contains the sensitive data it aims to protect. Organizations must treat it as a regulated asset: access controls limiting who can read raw step inputs and outputs, retention policies aligned with regulatory requirements, and potentially a two-tier structure that separates governance metadata (always retained, broadly accessible for audit) from step content (retained under stricter controls, accessible only for incident investigation).
\item \emph{Delegation provenance.} When a primary agent delegates to a sub-agent via a composite step, the sub-agent's execution produces its own path $P'$. What governance information from $P'$ should propagate back into the primary agent's state vector---and at what granularity---is an unsolved design problem that we discuss further in Section~\ref{sec:open}.
\end{enumerate}

\section{Implications for the EU AI Act}\label{sec:euaiact}

The development of this framework was motivated in part by the governance requirements of the EU AI Act \citep{euaiact2024}, whose obligations for high-risk AI systems take effect in August 2026. This section reflects on what the framework implies for organizations deploying agents under the Act. Applying its requirements to agents requires interpretation, and that interpretation is still being worked out by regulators and practitioners. What the framework provides is machinery: a precise structure for recording, evaluating, and enforcing governance decisions that can support whatever interpretation ultimately prevails. We list the most consequential articles:

\begin{enumerate}
\item \emph{Risk management throughout the lifecycle (Article 9).} The Act requires providers of high-risk AI systems to establish, implement, and maintain a risk management system throughout the entire lifecycle of the system. For agents, lifecycle risk management translates directly to the governance objective of Section~\ref{sec:engine}: the Policy Engine's continuous evaluation of $v_T$ against the organizational budget $B$ is a runtime instantiation of lifecycle risk management. The registration phase addresses deployment-time risk (documentation, integrity, scheduling); the per-step phase addresses operational risk; agent profiling closes the loop by feeding runtime observations back into design-time decisions. The Act's requirement that the risk management system be documented and repeatable is satisfied by the deterministic policy functions and the versioned audit trail: given the trail, any risk assessment decision can be reproduced.
\item \emph{Automatic logging (Article 12).} The Act requires that high-risk AI systems be designed to automatically generate logs of their operation, to the extent this is technically feasible. The audit log described in Section~\ref{sec:sysarch} directly addresses this requirement: every step, policy evaluation, score, decision, and outcome is recorded. Crucially, the log records not just what the agent did but what the governance system decided about it, including the policy version in force at the time. This constitutes active oversight documentation rather than passive execution recording. The privacy tension noted in Section~\ref{sec:challenges}---the log contains the data it aims to protect---is itself a compliance concern under the Act's data protection requirements and must be addressed in any compliant implementation.
\item \emph{Human oversight (Article 14).} The Act requires that high-risk AI systems be designed so that natural persons can effectively oversee them during use and intervene when necessary. A possible intervention (action) that pauses execution for human approval is a direct implementation of this requirement: for policies that require human approval before high-risk actions, the Policy Engine pauses execution, surfaces the relevant path context to a responsible person, and awaits their decision before proceeding. The override mechanism---a human approving or rejecting a proposed action, with the decision logged and the path updated accordingly---is precisely the ``meaningful human oversight'' the Act envisions. The practical caveat of Section~\ref{sec:boundary} applies here too: human oversight is only meaningful if the governance layer that decides when to invoke it is well-calibrated. Routing every agent action to a human reviewer satisfies the letter of Article 14 while defeating its purpose.
\item \emph{Transparency and documentation (Articles 13 and 16).} The Act requires that high-risk AI systems be sufficiently transparent and that providers maintain technical documentation. The policies requiring documented purpose, risk classification, and ownership as preconditions for agent registration address Articles 13 and 16 directly: an agent that cannot pass the documentation policy cannot run. The audit log provides the technical documentation of operational behavior. Together, these implement a documentation regime that is enforced rather than merely required.
\item \emph{Accuracy, robustness, and cybersecurity (Article 15).} The Act requires that high-risk AI systems achieve appropriate levels of accuracy and robustness, including resilience against attempts to alter their behavior through adversarial manipulation. The prompt injection scenario of Section~\ref{sec:whydifferent} is precisely such an attempt: retrieved content that manipulates the agent into violating a policy. The PII predecessor requirement policy addresses this by requiring a classification step before any action on personal data, regardless of how the agent's context was constructed. More broadly, the runtime governance layer provides robustness against a class of adversarial inputs that design-time measures cannot anticipate, because those inputs arrive at runtime via retrieved context. The Act's authors likely did not anticipate that the compliance infrastructure itself would become an attack surface---a compromised Policy Engine could approve prohibited actions---but this is a genuine robustness concern for any production deployment.
\end{enumerate}

The Act's requirements map naturally onto the framework's components, which is not coincidental: the framework was developed with these requirements in view. What the framework cannot provide is the calibration that makes compliance claims credible: the translation of $B$ into a specific risk budget, the empirical validation of $\pi_j$ outputs as genuine probabilities, and the demonstration that $\delta$ is set appropriately for the organization's risk profile. These require operational data and regulatory engagement that go beyond what any framework can specify in advance.

\section{Discussion}\label{sec:discussion}

This paper has argued that the governance of AI agent fleets reduces to a single object: a policy function $\pi_j(A, P_i, s^*, \Sigma) \to [0,1]$ evaluated before each proposed action. The argument has three parts.

The first is diagnostic. Agents are different from prior automated systems in ways that matter for governance: their behavior is non-deterministic, their tool use is decided at runtime, their paths vary in length, they can modify themselves, and they interact with other agents in ways that create organizational compliance constraints no individual agent can resolve. These properties mean that the violations organizations care about---data exfiltration, information barrier breaches, unauthorized external communication---are properties of sequences of actions, not of individual actions in isolation. Existing mechanisms (prompting, access control, guardrails, content filtering, human approval) are each valuable and each insufficient for the same reason: none can express or enforce a policy whose violation condition depends on prior steps.

The second is formal. We defined the execution path as the sequence of stochastic, deterministic, and composite steps an agent takes on a task, and the policy function as a deterministic map from path, proposed action, agent identity, and shared governance state to a violation probability. We showed that prompting is not an instance of this function at all (it modifies path distributions without evaluating them), and that access control is a degenerate instance (it uses only agent identity and action type, ignoring path and context). Runtime evaluation is the general case. We introduced the Policy Engine as the organizational component that evaluates policies across the fleet, maintains shared governance state, and calibrates $\delta$ so that $\mathbb{E}[\sum v_T] \leq B$ as the governance objective. Per-step evaluation is the mechanism; fleet-level risk management is the objective. 

The third is practical. We showed how the abstract framework instantiates into concrete policies, a two-phase architecture (registration and per-step), a reference implementation, and a mapping to EU AI Act requirements. The framework is not a complete system; it is a precise specification of what a complete system must do, general enough to survive changes in the underlying technology and concrete enough to guide implementation today.

\subsection{Open problems}\label{sec:open}

The framework identifies several problems it does not solve, each of which is a substantive research and engineering challenge.

\begin{itemize}
\item \emph{Risk calibration.} The policy function $\pi_j$ outputs a value in $[0,1]$ interpreted as a violation probability. For this interpretation to support meaningful fleet-level governance, the outputs must be calibrated: a policy that returns $0.7$ should correspond to a genuinely $70\%$ likely violation, not an arbitrary severity score. Current implementations often treat policy outputs as severity indicators rather than calibrated probabilities. Calibrating $\pi_j$ from operational data requires labeled execution traces---paths annotated with ground-truth violation judgments---which are not yet available at the scale needed. Deriving appropriate thresholds for $\delta$ from an organizational risk budget $B$ faces the same data requirement. This is the central open problem: without calibration, the probabilistic framework is a useful approximation, not a rigorous guarantee.
\item \emph{Strategic circumvention.} The framework assumes that agents are not modeling the governance layer and planning paths to stay under thresholds while achieving prohibited outcomes. For current-generation agents, this is a reasonable assumption: they have no persistent self-model across tasks and no incentive structure that would lead them to strategically circumvent governance. But it is a fragile assumption. An agent with long-term memory, access to its own policy scores, and a task objective that conflicts with governance constraints could, in principle, learn to produce paths that individually score low but collectively achieve a prohibited outcome. Runtime governance as described here catches incidental violations; it cannot anticipate strategy. A promising direction is multi-step lookahead: extending $\pi_j$ to score not only $s^*$ but a simulated continuation of the path, evaluating the projected trajectory before permitting the current step. Whether this can be made robust without reintroducing model-based stochasticity into the policy function is an open question.
\item \emph{Completeness of enforced compliance.} Prospective interception is enforced under the assumption that the agent operates within the governed execution environment. Agents with code execution capabilities can violate this assumption by spawning processes, making raw API calls, or modifying their own execution context in ways the Policy Engine does not observe. Enforced compliance is therefore an architectural constraint on agent capabilities, not a proven invariant of the governance system. For high-risk deployments, limiting agent access to code execution---or sandboxing it within the governed environment---is a practical requirement that must be addressed at the infrastructure level, not the policy level. A related granularity concern arises when a stochastic step generates executable code that is then run as a deterministic step: the step boundary as defined in Section~\ref{sec:paths} may be too coarse to detect violations buried inside the generated script. This is a further instance of the completeness problem, resolved by treating code execution either as a composite step subject to its own governance, or via sub-step instrumentation.
\item \emph{Behavioral drift.} \citet{rath2026drift} demonstrate measurable degradation in multi-agent systems over extended interactions: agents whose behavior gradually shifts from their initial configuration in ways that individually appear within tolerance but cumulatively represent significant deviation. Per-step evaluation with fixed policies may not detect drift if each individual step scores below intervention thresholds. Agent profiling---tracking the distribution of $v_T$ across repeated executions---is the natural detection mechanism, but it requires a baseline to compare against and a criterion for when drift has become governance-relevant. Connecting per-step evaluation to long-run behavioral monitoring is an open design problem.
\item \emph{Delegation provenance.} When a primary agent delegates to a sub-agent via a composite step, the sub-agent produces its own path $P'$ under its own governance evaluation. What should propagate from $P'$ back into the primary agent's state vector is not obvious. At minimum, the maximum sensitivity level and any barrier tags activated during $P'$ should propagate, so that the primary agent's subsequent policies have access to relevant context. Whether $v_T^{(\text{sub})}$ should contribute to the primary agent's $v_T$, and how to account for sub-agent violations in the fleet-level budget $B$, depends on how the organization wants to attribute responsibility across agent hierarchies. This is both a technical and a governance question without a settled answer.
\item \emph{Generated-code completeness.} The step taxonomy treats a deterministic step as an atomic triple $(\tau, d_{\mathrm{in}}, d_{\mathrm{out}})$. When the output of a stochastic step is a script executed as a single deterministic step, violations inside the generated code are invisible to the policy function unless the execution environment decomposes the script into sub-steps. This is a specific instance of the completeness problem: governance at step granularity cannot detect violations at sub-step granularity. Sandboxed execution environments that expose individual system calls as governed steps are the natural mitigation, but their integration with the Policy Engine is an open engineering problem.
\item \textit{Fleet prioritization under budget exhaustion.}
The objective constrains $\mathbb{E}[\mathrm{violations}(t)] \le B$ but does not specify which tasks to terminate when the budget is reached. Any prioritization rule—by task value, agent risk class, or arrival order—is a policy choice outside the framework. Organizations deploying at scale will need an explicit scheduling policy for this case; the framework provides the constraint and the monitoring signal, not the optimization logic.
\item \emph{Policy interaction at scale.} The composition rule $v_i = 1 - \prod_j(1 - \pi_j)$ means that adding policies increases $v_i$ even if each individual policy has low scores. In a large policy set, the aggregate step-level violation probability can be driven high by many individually low-scoring policies simultaneously firing on the same action, potentially blocking actions that no individual policy would block on its own. Testing policies for interaction effects requires simulation over realistic path distributions, which is expensive and requires representative traces. Organizations deploying large policy sets will need tooling to detect and manage policy interactions that does not yet exist.
\end{itemize}

\subsection{Next steps}\label{sec:nextsteps}

The most immediate need is empirical validation: deploying the framework in production environments, measuring per-step overhead, and beginning the process of calibrating $\pi_j$ outputs against ground-truth violation judgments. Without operational data, the probabilistic framework rests on untested assumptions. Calibration requires not only labeled traces but agreement on what constitutes a violation---which for many policies is a regulatory and organizational question as much as a technical one.

A natural formal extension is to connect the fleet-level framework to per-agent guarantees of the kind developed by \citet{bhardwaj2026abc}. Their ABC provide probabilistic compliance guarantees for individual agents under specific behavioral assumptions; the fleet-level framework provides the organizational context within which those guarantees must hold. Combining the two would yield a framework with per-agent formal guarantees and fleet-level risk management---a more complete governance architecture than either provides alone.

Finally, the framework as presented governs agents within a single organization. Many enterprise deployments involve agents that span organizational boundaries---supply chain agents, inter-organizational automation, third-party agent services integrated into internal workflows. Extending the shared governance state $\Sigma$ and the fleet-level objective to multi-organization settings raises questions of trust, liability, and information sharing that the current framework does not address. This is a longer-term research direction, but one with significant practical importance as agent deployment matures.


\bibliographystyle{plainnat}
\bibliography{references}

\end{document}